%% file: ijcai22-multiauthor.tex
\documentclass{article}
\usepackage{ijcai22}

\usepackage{times}

\usepackage{soul}
\usepackage{url}
\usepackage[hidelinks]{hyperref}
\usepackage[utf8]{inputenc}
\usepackage[small]{caption}
\usepackage{graphicx}
\usepackage{amsmath}
\usepackage{booktabs}
\usepackage{multirow}
\usepackage{lscape}
\usepackage{amssymb}
\usepackage{lscape}
\usepackage[table]{xcolor}
\usepackage{array}
\usepackage{enumitem}
\usepackage{tikz}

\title{A Survey of Machine Narrative Reading Comprehension Assessments}

\author{
Yisi Sang$^1$\and
Xiangyang Mou$^2$ \and 
Jing Li$^3$\and
Jeffrey Stanton$^{1*}$ \and
Mo Yu$^4$\footnote{Contact Authors.}
\\
\affiliations
$^1$Syracuse University\\
$^2$Rensselaer Polytechnic Institute\\
$^3$New Jersey Institute of Technology\\
$^4$WeChat AI, Tencent Inc.\\
\emails
\{yisang, jmstanto\}@syr.edu,
moux4@rpi.edu,
jingli@njit.edu,
moyumyu@tencent.com
}

\usepackage[prependcaption,textsize=tiny]{todonotes}

\begin{document}

\maketitle

\begin{abstract}
As the body of research on machine narrative comprehension grows, there is a critical need for consideration of performance assessment strategies as well as the depth and scope of different benchmark tasks. Based on narrative theories, reading comprehension theories, as well as existing machine narrative reading comprehension tasks and datasets, we propose a typology that captures the main similarities and differences among assessment tasks; and discuss the implications of our typology for new task design and the challenges of narrative reading comprehension.

\end{abstract}

\input{1-introduction}

\input{Narrative_Expository}

\input{3-Task-Formats}

\input{2-Narrative-Element}

\input{5-taxonomy}

\input{discussion-conclusion}


\bibliographystyle{named}
\bibliography{reference}

\end{document}

%% file: 1-introduction.tex
\section{Introduction}\label{sec:introduction}

Expository texts that provide facts and information about a topic
and narrative texts that present a story are the two main text genres for reading comprehension.
From the perspective of cognitive and narrative theories, understanding narratives is a complex process that requires the development of multiple capabilities at the same time, such as story grammar, theory of mind, and perspective-taking~\cite{paris2003assessing}. In NLP research, people have developed linguistic resources and tools for analyzing narrative texts~\cite{bamman2020litbank}, and evaluation benchmarks on various high-level narrative understanding tasks such as event relation identification~\cite{glavavs2014hieve}, question answering~\cite{kovcisky2018narrativeqa} and story summarization~\cite{chen2021summscreen}.

Previous survey papers on machine reading comprehension (MRC) have covered methods, trends~\cite{liu2019neural},
and datasets~\cite{dzendzik2021english}. However, the state of the art was such that these papers did not need to distinguish between expository texts and narrative texts, which each have unique requirements in comprehension. Thus, the uniqueness of narrative comprehension and the complexity of its evaluation have not been reflected by previous surveys. Over the past few years, the natural language processing (NLP) community has made rapid progress in improving the performance of neural models for machine reading comprehension (MRC). Combining expository and narrative sources in previous work under the broad umbrella terms of machine narrative reading comprehension has arguably slowed progress in evaluative methods. As each of these tasks aims to assess a specific perspective of narrative understanding, we believe that there is much to gain by studying how they are related and how they are different. 

To help bring together the various approaches to assessment and to differentiate the depth and scope of their evaluation of narrative understanding, we propose a typology that synthesizes these different assessments. In our typology we argue that the differences between assessments of machine narrative reading comprehension can be reduced to two informative dimensions:
\begin{itemize}[leftmargin=*]
\setlength\itemsep{0em}
\item Local versus global narrative representation (i.e., the extent of the text stream over which the reader needs to link narrative elements)
\item Extent of narrative elements extracted by the comprehension model
\end{itemize}

These two dimensions can be used to category the existing assessments of machine narrative reading comprehension. Our goal is to clarify the differences and similarities between assessments of narrative machine reading comprehension to help researchers select appropriate assessment tasks to evaluate models of narrative reading comprehension and to shed light on emerging and often overlooked challenges when building machine narrative comprehension tasks.

We organize our paper in the following way: first, we differentiate narrative text and expository text and review the difficulties in understanding narrative texts; second, we illustrate the theoretical foundations of a typology by summarizing the fundamental elements of narratives, scopes of comprehension, and types of existing tasks; next, we review existing narrative machine comprehension datasets; finally, we describe the typology and discuss the research opportunities within each dimension.

%% file: Narrative_Expository.tex
\section{Background: Narrative vs. Expository}

In this section we review the difference between narrative and expository texts, showing the uniqueness of narrative comprehension and by extension the value of this survey.

Narrative and expository texts, as two different discourse genres, have different communicative goals and functions. They differ in their principles of linguistic expression and organization.
According to Brewer’s genre classification system~\cite{brewer2017literary}, a narrative text is defined as one in which events that are related causally or thematically occur chronologically. In contrast, expository texts are defined as texts that describe a system or event in terms of its processing or structure. 
Narratives tend to be agent-oriented with a focus on characters, their actions, and their motivations. Narratives express the development of events within a temporal framework.
Expository texts, on the other hand, are topic-oriented; they focus on one or more concepts and express the development of ideas, assertions, and arguments in terms of their logical interrelationships~\cite{britton1994understanding}.

The role of events in narrative is significantly different from their role in fact-based expository texts of real-world events. Stories are usually longer and have more complicated narrative structures than expository texts, both locally and globally. Furthermore, stories are a creative endeavor in which the causality of real-world events is not hard-coded into narrative event sequences.

Narrative and expository texts also imply different perspectives on the nature of understanding. Human readers process narratives in order to create explanation-based coherence. Narrative processing is concerned with understanding the organization of events in the story ~\cite{wolfe2010processing}. 
Readers often make inferences based on general world knowledge to explain how aims, events, actions, and outcomes in stories are related.
These inferences represent links between narrative events, connections to readers' existing knowledge, and predictions about what will happen next~\cite{wolfe2010processing}. Expository processing is more concerned with the activation and integration of relevant prior knowledge into the discourse representation. The understanding of expository texts is often characterized by readers' attempts to construct a coherent representation of concepts extracted from the text content.

%% file: 3-Task-Formats.tex
\section{Background: A Survey of Task Formats for Assessments of Narrative Compression}

Because narrative reading comprehension is a special category of MRC, it has also traditionally been assessed using traditional MRC tasks. However, text genre plays an important role in comprehension. Most traditional MRC tasks focus on expository texts, so these forms of assessment may not be appropriate for assessing reading comprehension of narrative texts. In this section we summarize the traditional MRC task formats that have been applied to narrative comprehension, and analyze the advantages and disadvantages of its application to reading comprehension of narrative texts. 

\paragraph{Cloze Test} 
takes a snippet of the original text with some pieces (usually entities) masked as blanks, with the goal of filling these blanks from a list of candidates. 
Examples of cloze tests for narrative comprehension assessments include BookTest~\cite{bajgar2016embracing},
and~\cite{ma2018challenging}. However, when building on short snippets, the cloze tests is known to prone to mostly local inference but not much reasoning and commonsense knowledge, as pointed by studies in the NLP community suggested
~\cite{chen2016thorough}. 

\paragraph{Question Answering (QA)} is widely considered to be a generalized task format for MRC. However, given the challenge of having human annotate large-scale questions, creating a QA dataset to accurately assess certain reading skills can be quite difficult. Lengthy narratives make this problem even more difficult, as good assessment questions, especially those that require global information, usually require crowd workers to read and achieve a thorough understanding of long stories. This difficulty makes existing benchmarks mainly have questions on local snippets~\cite{yang2019friendsqa}, short stories~\cite{richardson2013mctest,xu2022fantastic}, with the only exception of~\cite{kovcisky2018narrativeqa}. 
Therefore, more attention could beneficially be paid to carefully designed task formats by experts beyond QA, for efficient assessment of narrative reading skills.

\paragraph{Summarization} has a main focus on plot line understanding. There has been considerable recent interest in evaluating a model's understanding of stories via summarization, e.g., NovelChapters~\cite{ladhak2020exploring}, BookSum~\cite{kryscinski2021booksum} and ScreenSum~\cite{chen2021summscreen}. Intuitively, summarization requires a deep understanding of the global information of a story, to enable generation of story summaries. These tasks provide difficult challenges to existing machine reading models. However, summarization tasks also have a significant drawback insofar as there are many factors beyond reading skills involved in generating a good summary, such as generating long narrative texts. As a result, summarization is not a pure measure of reading comprehension.

\paragraph{Fundamental Language Annotation Tasks} 
refers to standard NLP tasks from syntactic to semantic analysis that provide bases for narrative understanding as well. Standard NLP tasks are hence extended to the narrative domain, including part-of-speech (POS) tagging and named entity recognition (NER) about location, time, and character names, event detection, and coreference resolution~\cite{bamman2019annotated}.

\paragraph{Story-Level Classification} 
refers to a wide range of tasks with the format of classification, which requires the information collected across the whole story. One example is the prediction of characters' personality types by reading the original stories~\cite{flekova2015personality}.

%% file: 2-Narrative-Element.tex
\section{A Survey of Theories of Narrative Elements}
\label{sec: element}

Narrative stories contain consistent structural elements, but these elements have been divided and defined in a variety of ways. In this section we review key theories that describe and analyze narrative elements.

\paragraph{Theories in Social Science} According to research on situation models, humans pay attention largely to spatial and event related information in a narrative. In a typical story, each event is indexed according to its time period, its location, the main characters it involves, its causal relationship to earlier events, and its relevance to the protagonist's goals. The reader then determines whether an index must be updated for the next encountered story event according to any of these situational dimensions~\cite{zwaan1995construction}.

Story structure theories explore the functional elements of a narrative. For example, ``story scheme'' refers to a set of expectations about the internal structure of a story that facilitates encoding and retrieval. Syntactic categories include setting, event, change-of-state, emotion, desire, action, plan and subgoal~\cite{rumelhart1975notes}. Gordon Bower proposed  rules to clarify the structure of stories and the process of human understanding. The first rule defines that a story consists of a setting, a theme, a plot, and a resolution, and they usually appear in that order. The second rule is that the setting consists of the characters, as well as the place and time of the story. The third rule is that the theme of a story consists of the main objectives of the main characters~\cite{guthrie1977research}.

In education research, story elements including main character, setting (i.e., time and location), 
problem, attempted solution,
and ultimate solution are often used to assess students' narrative comprehension~\cite{garner2004transfer}. \cite{paris2003assessing} classified story elements into implicit and explicit~. Specifically, five explicit (i.e., setting, character, initiating event, problem, and outcome resolution) and five implicit (i.e., character feelings, character dialogue, causal inference, prediction, and theme) text relations. Experiments showed that children develop schemata about the settings, actions, and events described in narratives~\cite{cote1998students}

\paragraph{NLP Theories}

The NLP community draws on a range of social science theoretical perspectives on narrative to form evaluation tasks, so it may have value to place these perspectives within an organized theoretical structure that can be applied to these practical machine evaluation tasks.

Most NLP studies mainly follow only the event-centric perspective and highlight causal chains, plans, and goals as important components of comprehending stories.
But recent works have started to consider a more comprehensive view of narrative understanding.
\cite{dunietz2020test} suggested four overlapping clusters of questions for narrative comprehension, extending from events to agents' reactions to the events, which correspond to the four elements highlighted in~\cite{zwaan1995construction}.
The question templates include the three common types in previous NLP studies such as spatial questions, temporal questions and causal questions, plus an additional type of motivational questions such as, ``\emph{how do agents’ beliefs, desires, and emotions lead to their actions}''.

\cite{piper2021narrative} linked computational work in NLP to narrative theoretical frameworks and proposed a working definition of narrativity. The definition emphasized the audience interaction between narrative features and audience interactions with feature level interactions. They proposed eight elements that must be present in order to form a narrative: teller, mode of telling, recipient, situation, agent, one or more sequential actions, potential object, spatial location, temporal specification, and rationale.
As will be shown in Section~\ref{sec:typology}, their defined elements have overlaps to our typology. However, their work aims at a general-purposed definition of narratives, while ours focuses on the narrative story structures; thus we cover several important elements missed in their work.
Also, similar to the reviewed theories from social science, their defined elements follow a different granularity, thus is less well-aligned to assessment tasks in the NLP field.

%% file: 5-taxonomy.tex
\section{Our Proposed Assessment Typology}
\label{sec:typology}

\begin{table}[]

\resizebox{\columnwidth}{!}{%
\centering
\begin{tabular}{lcclc}
 &
  \multicolumn{1}{c}{\textbf{\begin{tabular}[c]{@{}l@{}}Event\end{tabular}}} &
  \multicolumn{1}{c}{\textbf{Character}} &
  \multicolumn{1}{c}{\textbf{Setting}} &
  \multicolumn{1}{c}{\textbf{\begin{tabular}[c]{@{}l@{}}Functional \\ Structure\end{tabular}}} \\ \cline{2-5} 
\multicolumn{1}{c|}{\rotatebox[origin=c]{90}{\textbf{Global}}} &
  \multicolumn{1}{c|}{\begin{tabular}[c]{@{}c@{}}\textbf{NarrativeQA}\\ \scriptsize{~\cite{kovcisky2018narrativeqa}}\\ event structure\end{tabular}} &
  \multicolumn{1}{c|}{\begin{tabular}[c]{@{}c@{}}\textbf{TVSG}\\\scriptsize{~\cite{sang2022tvshowguess}}\\ character persona \\ understanding\end{tabular}} &
  \multicolumn{1}{l|}{\cellcolor[HTML]{C0C0C0}} &
  \multicolumn{1}{c|}{\begin{tabular}[c]{@{}c@{}}\scriptsize{~\cite{papalampidi2020screenplay}}\\ Screenplay summarization\end{tabular}} \\ \cline{2-5} 
\multicolumn{1}{c|}{\rotatebox[origin=c]{90}{\textbf{Local}}} &
  \multicolumn{1}{c|}{\begin{tabular}[c]{@{}c@{}}\textbf{ESTER}\\\scriptsize{~\cite{han2021ester}}\\ event relation \\ understanding\end{tabular}} &
  \multicolumn{1}{c|}{\begin{tabular}[c]{@{}c@{}}\textbf{LiSCU}\\ \scriptsize{~\cite{brahman2021let}}\\ character identification \\over summaritive texts \end{tabular}} &
  \multicolumn{1}{c|}{\begin{tabular}[c]{@{}c@{}}\textbf{LitBank}\\ \scriptsize{\cite{bamman2020litbank}}\\ location NER\end{tabular}} &

  \multicolumn{1}{c|}{\begin{tabular}[c]{@{}c@{}}\textbf{TRIPOD}\\\scriptsize{~\cite{papalampidi2019movie}}\\ turning point detection\end{tabular}} \\ \cline{2-5} 
\end{tabular}
}
\caption{\small A typology for evaluating narrative machine understanding }
\label{tab: typology}
\end{table}

We synthesize the existing assessment tasks on narrative reading comprehension into a two-dimentional typology that considers (i) the scope of texts required to solve the tasks; and (ii) the target narrative elements to be assessed. Compared to the categorizations discussed in Section \ref{sec: element}, our typology is tailored for NLP research and makes a clearer distinction among NLP tasks than prior work. Specifically, the types in our typology can be well aligned to the focuses of existing NLP datasets. Table~\ref{tab: typology} illustrates our typology and the representative tasks for each category.

\subsection{Meaning Representation Scope of a Narrative}

Forming adequate representations of narrative elements and the development of structural knowledge of the relations among elements are crucial for successful comprehension.
One of the most well-established reading comprehension models, the Construction-Integration (CI) model~\cite{kintsch1988role}, illustrates that a reader's representation can be based upon microstructure or macrostructure. Microstructure is driven by the local structure of the narrative (e.g., individual scenes), while macrostructure is driven by the global or hierarchical structure of the narrative (e.g., the entire story).
The microstructure includes the reader's local inferences, but 
does not connect larger scale elements of the narrative. The hierarchical macrostructure 
requires the reader to infer the global organization of the narrative by connecting multiple microstructure elements~\cite{mcnamara2009toward}. 

Past research has indicated the existence of a significant amount of evaluation of local representation such as recognizing relations between characters~\cite{chen2016character} and narrative scene detection~\cite{delmonte2017semantically}. 
However, only a few assessments has been developed that require global representation. These assessments include direct evaluation of a skill through a specific task, such as predicting personality types to assess character understanding, and indirect evaluation of a skill embedded in a task, such as identifying characters based on their dialogue, which requires an implicit theory of mind. The rows of table\ref{tab: typology} shows tasks requiring global or local representation .

\subsection{Target Narrative Element}
\label{ssec:typology_element}
Based on narrative theories, in this section we classify the target of existing machine narrative reading comprehension assessments into fundamental narrative elements: event, character, setting, and functional structure. 

\paragraph{Event}
In narrative theories, an event is actually an implicit element. An ``event'' implies the occurrence of a transformation, while the more atomic idea of ``activity'' entails agents; when agents act, they must have motivations and be attempting to solve problems. Motivations may or may not be made explicit in the narrative. Agents may face challenges or they may be involved in some type of conflict~\cite{ryan2007toward}. This means that when one discusses the events based on these theories they are actually talking about a structure containing several elements with different components. 

In NLP, however, the scope of narrative event is often construed much more narrowly than the implicit event - typically more on the level of an activity. For example, an event has been defined as ``a tuple of a frame (most simply a verb) and its participants''~\cite{chambers2008unsupervised}. 
This definition of events as (verb) frames usually lies on a lower-level compared to the customary definition people usually refer to in daily lives.
The latter is usually a sequence of the former ``events'' under the same theme.
While the NLP events are usually on the sentence- or paragraph-level,  the customary usage of events can refer to a scene or even a whole plotline. 

In this survey, in order to propose a typology that is more applicable to a range of assessment tasks, when we refer to events we follow both the material. Therefore the scope of the event element include a hierarchy from people's actions, the various changes in nature, to the customary events; and moreover the relations and structures of the above events. The first column of Table \ref{tab: typology} shows cases of event related assessments.

\paragraph{Character}
Characters are agents such as people, animals, and other creatures in a story. A character-centered perspective seeks to understand the characters that make up the story such as understanding the characters' roles, goals, relationships, emotions, and personality. Character identification and personality prediction are character-related assessments.

\paragraph{Setting}
Previous work has usually defined setting as the physical universe in which action takes place~\cite{piper2021narrative}. However, in narrative, settings can also be historical and contemporary contexts, which are used to set the mood and shape the subjective atmosphere. Subjective atmosphere is not a one-time description, but is perceived and understood by the reader through various descriptive elements throughout the text.

\paragraph{Functional Structure}
Functional structure refers to an abstract representation of the different contributions of higher level aspects of a narrative to its intended function~\cite{zan1983toward}. Functional structure is conceptually similar to a grammar that focuses on function rather than content.
The primary distinction between functional structure and scripts is that the scripts contain events from the narrative whereas functional structure are made up of phases in a story arc.
Turning point detection is an example of understanding functional structure.

\begin{table*}[!ht]
\small
\centering
\resizebox{0.96\textwidth}{!}{%
\begin{tabular}{ccccccc}
\toprule
\multirow{2}{*}{\textbf{Dataset}}& \multirow{2}{*}{\textbf{Task Format}}& \multirow{2}{*}{\textbf{Narrative Source}}&  \multicolumn{4}{c}{\textbf{Targeted Story Elements}}\\ 
& & &  \textbf{Event} & \textbf{Character} & \textbf{Setting}& \textbf{Functional Structure}\\
\midrule

{\begin{tabular}[c]{@{}c@{}}\textbf{MCTest} \scriptsize{~\cite{richardson2013mctest}}\end{tabular}} & multi-choice & children stories &\checkmark  &\checkmark &    \\
{\begin{tabular}[c]{@{}c@{}}\textbf{CBT} \scriptsize{~\cite{hill2015goldilocks}}\end{tabular}} & cloze test & children stories   &\checkmark  & &    \\
{\begin{tabular}[c]{@{}c@{}}\textbf{LAMBADA}\scriptsize{~\cite{paperno2016lambada}}\end{tabular}} & language model & literature  &\checkmark  & &    \\
{\begin{tabular}[c]{@{}c@{}}\textbf{literary events} \scriptsize{~\cite{sims2019literary}}\end{tabular} } & event trigger detection &literature
&\checkmark  & & &   \\
{\begin{tabular}[c]{@{}c@{}}\textbf{HiEve} \scriptsize{~\cite{glavavs2014hieve}}\end{tabular} } & event relation detection &news stories
&\checkmark  & & &   \\
{\begin{tabular}[c]{@{}c@{}}\textbf{TORQUE}\scriptsize{~\cite{ning2020torque}}\end{tabular} } & event relation detection &news stories
&\checkmark  & & &   \\
{\begin{tabular}[c]{@{}c@{}}\textbf{TellMeWhy} \scriptsize{~\cite{lal2021tellmewhy}}\end{tabular}} & multi-choice & short fictions &\checkmark  & &   \\
{\begin{tabular}[c]{@{}c@{}} \textbf{MCScript}\scriptsize{~\cite{ostermann2018mcscript}}\end{tabular}} & multi-choice & daily narratives &\checkmark &  &    \\
{\begin{tabular}[l]{@{}l@{}}\textbf{ROCStories}\\\quad \scriptsize{~\cite{mostafazadeh2016corpus}}\end{tabular}} & multi-choice & short stories&  \checkmark &    \\
{\begin{tabular}[c]{@{}c@{}}\textbf{NarrativeQA}\scriptsize{~\cite{kovcisky2018narrativeqa}}\end{tabular}} & free-answering QA & movie scripts, literature& \checkmark  &\checkmark &\checkmark    \\
{\begin{tabular}[c]{@{}c@{}}\textbf{FriendsQA} \scriptsize{~\cite{yang2019friendsqa}}\end{tabular}} & extractive QA & TV show scripts &\checkmark  &\checkmark &\checkmark    \\
{\begin{tabular}[c]{@{}c@{}}\textbf{NovelChapters} \scriptsize{~\cite{ladhak2020exploring}} \textbf{/}\\ \textbf{BookSum} \scriptsize{~\cite{kryscinski2021booksum}}\end{tabular}} & summarization & literature &\checkmark  & &    \\
{\begin{tabular}[c]{@{}c@{}}\textbf{SumScreen} \scriptsize{~\cite{chen2021summscreen}}\end{tabular}} & summarization & TV show scripts &\checkmark  & &    \\
\textbf{\cite{flekova2015personality}} & classification & literature & &\checkmark &    \\
{\begin{tabular}[c]{@{}c@{}}\textbf{\cite{chen2016character} /} \\ \textbf{~\cite{chen2017robust}}\end{tabular}} & coref resolution & TV show scripts & &\checkmark &    \\
{\begin{tabular}[c]{@{}c@{}}\textbf{LiSCU} \scriptsize{~\cite{brahman2021let}}\end{tabular} } & cloze test & {\begin{tabular}[c]{@{}c@{}}paired (literature,\\ character) summaries\end{tabular}} &  &\checkmark & &   \\

{\begin{tabular}[c]{@{}c@{}}\textbf{\cite{massey2015annotating}}\end{tabular} } & relation detection &literature &  &\checkmark & &   \\
{\begin{tabular}[c]{@{}c@{}}\textbf{TVSG} \scriptsize{~\cite{sang2022tvshowguess}}\end{tabular} } & character guessing &TV show scripts &  &\checkmark & &   \\
\textbf{\cite{bamman2019annotated}} & coref resolution & literature & & &\checkmark    \\

{\begin{tabular}[c]{@{}l@{}}\textbf{TRIPOD}\\ \quad \scriptsize{~\cite{papalampidi2019movie}}\end{tabular}} &turning
points identification & movie scripts &  & & &\checkmark   \\

{\begin{tabular}[c]{@{}c@{}}\textbf{CompRes} \scriptsize{~\cite{levi2020compres}}\end{tabular} } & classification  & news stories  &  & & &\checkmark   \\

{\begin{tabular}[c]{@{}c@{}}\textbf{\cite{ouyang2014towards}}\end{tabular} } & classification  &personal experience  &  & & &\checkmark   \\

\bottomrule
\end{tabular}
}
\caption{\small{Popular evaluation datasets of machine narrative reading comprehension.}}
\label{tab:existing_ds2}
\end{table*}

\section{Organizing Assessments in Our Typology}
In this section we review existing assessment benchmarks according to our typology. The datasets are organized according to the narrative elements and summarized in Table~\ref{tab:existing_ds2}, with the categories along the two dimensions discussed in texts.

\subsection{Event-Centric}
Historically, the representation and identification of events and their participants in NLP have focused on the domain of news, including early evaluation campaigns such as seminal datasets ACE2005~\cite{walker2006ace} and other resources that necessitate event identification as a prerequisite for other tasks such as temporal ordering or factuality judgments. In narrative understanding event-centric research covers a broader topics.

\paragraph{Event Detection}
The dataset literary events~\cite{sims2019literary} identifies events that are depicted as actually happening. In other words, events that are asserted to be real. Their events includes activities, achievements, accomplishments, and changes of state as being events. Event triggers in this dataset is limited to verbs, adjectives, and nouns. 
Like the standard event detection tasks, these problems can usually be solved in local contexts, even on the sentence-level.

\paragraph{Event Relation}
The majority of new event-centric tasks in the NLP field focus on the prediction of a relation between two events, with both events provided and described in the narrative texts. 
The relationships considered include the causal and conditional relationships~\cite{mirza2014analysis,lal2021tellmewhy}; temporal relationships, which define the relationship between two events in terms of time, or between one event and a specific time point, such as tomorrow, as covered by the recent TORQUE~\cite{ning2020torque} dataset. 

Another relationship is the inclusiveness between a main event and one of its sub-events.
Note that this is different from the larger scope of ``events'' as discussed in Section~\ref{ssec:typology_element} and will be surveyed in the \emph{Customary Event Hierarchy} section, because the main events with larger scopes appear in the texts as well.
HiEve~\cite{glavavs2014hieve} is one such example on news stories. It represents the stories as event hierarchies — directed acyclic graphs (DAGs) of event mentions with edges denoting spatiotemporal confinement between events. The relationship of spatiotemporal confinement shows that one event is a component of another.

Finally, some datasets include multiple types of relationships. For example, RED~\cite{o2016richer} annotates causal and sub-event relations jointly. ESTER~\cite{han2021ester} proposed five types of event semantic relations: causal, sub-event, coreference, conditional and counterfactual. Event schema was proposed to learn high-level representations of complex events and their entity roles from unlabeled narrative text~\cite{chambers2013event}. As all these datasets assume the appearance of event mentions in local contexts, the relationships can always be identified with local inference. 

\paragraph{Event Precedence}

Several studies have looked at constructing sequences of events and modeling the linear order of occurrences. 
The trend of using scripts in narrative understanding started by the proposing of narrative event chain~\cite{chambers2008unsupervised}. 
The authors proposed an assessment called the narrative cloze test, designed to predict the absence of an event based on all other events in the script. Later the scope of event chains was expended by jointly learning event relations and their participants from unlabeled corpora.
\cite{ostermann2018mcscript} is another dataset assessing the understanding of script knowledge in narrative, with the task format of question answering.
Another task that belongs this category is story cloze, e.g., ROCStories~\cite{mostafazadeh2016corpus}, which requires to choose the correct ending for a story from the given endings.

\paragraph{Customary Event Hierarchy}
A largely ignored type of reading comprehension skill is the understanding of the customary events, which usually correspond a whole scene or plotline of the local NLP events.
The problem is thus recognizing the themes of these ``large'' events, as well as identifying their inner structures, i.e., how they are constructed from the ``small'' NLP events.

On this direction, the available assessments are limited: \cite{delmonte2017semantically} identified narremes which is the smallest unit of narrative structure. ~\cite{mikhalkova2020modelling} annotated main components of a storyline.
However, some general-purposed assessment tasks consist of small portions of such problems.
For example, the NarrativeQA dataset~\cite{kovcisky2018narrativeqa} may question the information of an event that has a scope across multiple paragraphs, as analyzed by \cite{mou2021narrative}.
The summarization tasks~\cite{ladhak2020exploring,kryscinski2021booksum} are another good example of this, since generating a chapter-level summary naturally requires to understand the event hierarchy and describe the upper-level events in concise texts.

\subsection{Character-Centric}

The task of coreference resolution for story characters~\cite{chen2016character,chen2017robust} focuses on identifying the characters mentioned in multiparty conversations. The goal of these tasks is to resolve the coreference of pronouns and character-indicating nominals (\emph{e.g.}, \emph{you} and \emph{Mom}) in dialogues of the character names that appear in the local context. It also covers linking a named entity (\emph{e.g.}, \emph{Ross}) to the character.
LiSCU~\cite{brahman2021let} is a dataset that contains summaries of literary works as well as summaries of the characters that appear in them.
The authors propose two tasks: character recognition as a cloze test and the generation of character descriptions. Both the aforementioned datasets require the understanding of characters' ``facts'' (i.e., their participated events over short spans), thus can be mainly resolved within local contexts. 

There are also tasks encouraging understanding characters in global contexts.
Inter-character relationship is a tradition for understanding narrative characters which is related to social network theories. \cite{massey2015annotating} created a dataset of manually annotated relationships between characters in literary texts.
Another character-centric task is to guessing characters by reading the stories~\cite{sang2022tvshowguess}. The task requires to comprehend the original long stories that contain the character's verbal and non-verbal narratives; hence needs a global representation of the narrative.

\subsection{Setting-Centric}
Existing assessment of understanding setting in narratives mainly focus on the time and place of a story. It often answers the questions about when and where. Modeling settings naturally requires the identification of locations. named entity recognition of locations is the typical task for psychical setting related tasks.
~\cite{bamman2019annotated} covered instances that related to location-related NER and coreference resolution in long documents. 
However, for the understanding of the more challenging setting cases such as historical and contemporary backgrounds, which require global representation in the narrative, there is no benchmark available.

\subsection{Functional-Structure-Centric}
Functional structure focuses on the functions of narrative fragments.
There are two line of functional structure research. For short narratives, based on William Labov’s theory of narrative analysis~\cite{labov1997narrative},~\cite{ouyang2014towards} detected complicating actions, CompRes~\cite{levi2020compres} identified complication, resolution, and Success in news articles, ~\cite{saldias2020exploring} disentangled narrative clause types. For long narratives, TRIPOD~\cite{papalampidi2019movie} analyzed plot structure by identifying ``turning point''.
The existing benchmarks examine the global representation of narrative structures, however, the hierarchical functions that facilitate the building of event-scene-plot-narrative still need further exploration.

%% file: discussion-conclusion.tex
\section{Discussion}

\subsection{Implication of the Typology}
Our survey suggests future improvements: First, while many datasets exist on event-centric assessments, the tasks for the other elements are relatively limited, especially for setting-centric and functional-structure-centric assessments.
It will be helpful to conduct comprehensive analysis of which narrative elements should be considered in the specific tasks and develop assessments accordingly.

Second, even for existing benchmarks, it is important to conduct analyses of which sub-tasks and reading comprehension skills present the greatest challenges.
Additionally, it is helpful to highlight overlooked distinctions of assessment used in existing narrative understanding tasks. 
In many narrative assessments the scope of meaning representation and types of inference are not well differentiated. 
For example, although both character name linking and character identification are character-centric narrative understanding, character identification requires pragmatic inference and global representation, character name linking usually requires propositional inference and local representation.

Finally, in narrative comprehension, the understanding module of different elements could interact with each other, which calls for assessments of joint understanding. For example, understanding characters and event can be jointly used for understanding the progression of narratives~\cite{phelan1989reading}.
By comparison, existing benchmarks that cover multiple elements in Table~\ref{tab:existing_ds2} usually have individual elements assessed by disjoint sub-sets of instances.

\subsection{Current and Future Challenges}

Narrative texts such as novels and even most short stories, are substantially longer than texts studied in conventional machine comprehension tasks and have more complicated narrative structures both locally and globally. Although there are some attempts to address these challenge, methods to accurately and efficiently handle long narrative input data remains a challenge for narrative comprehension.
To encourage machine narrative reading comprehension, more carefully designed tasks that require the global inference are helpful.

Additionally, a necessary step toward narrative understanding is pragmatic inference which is the interpretive process through which readers must reconcile the differences between literal and intended meaning of texts.
The incorporation of commonsense knowledge is one way to fill this gap. 
For example, the script knowledge~\cite{shank1977scripts} can help models to complete the omitted sub-processes of a main event depicted in a story and social commonsense, e.g.,~\cite{rashkin2018event2mind}, can help to reveal people's stereotypical intentions beneath the textual descriptions of their actions and dialogues.
Though specific commonsense inference tasks have been designed to encourage the study of such knowledge, incorporating it into the end tasks of machine narrative reading comprehension remains challenging.

Moreover, the expressions of narrative texts, such as argument, lyricism, and illustration; the narrative sequence, such as flashback, interpolation, and supplementary narrative; the viewpoint of the narration, such as first person, second person, and third person; the expressions of narrative texts, such as symbolism as well as desire to raise and lower; discursive forms such as dialogue and monologue may influence the difficulty of the task. A more detailed analysis of the input narrative text would help in the design of the task.

Furthermore, despite some reading skills and narrative sources that have been covered by existing datasets, there are still some missing assessments.
One example is the understanding of the intentions of speakers, which play an essential role in stories, especially in dramatic scripts.
However, there exist few assessment datasets on understanding the intentions in dialogues and how they would push forward the story progressions.
This missing assessment limits the models from dealing with dialogues and non-dialogues in a different but cooperative manner.

Finally, text genre and inference types could be different dimensions of reading comprehension. In this survey we did not discuss the types of inference in depth, but we noticed that there are many more tasks available for propositional inferences than for pragmatic inferences. 
Pragmatic is an important dimension for both narrative and expository texts.
While in narrative texts, it is more often authors hide deep meanings under the surface.
More pragmatic inference tasks need to be designed in the future.

\section{Conclusion}
We present a typology that synthesizes the different assessment tasks in machine narrative reading comprehension. By making connections between cognitive theories, narrative theories and existing research in NLP, we hope to bring together findings in these different areas and to clarify the key aspects, overlooked distinctions and suggest major research challenges that will help drive the empirical study of machine narrative reading comprehension forward. Rather than attempting to solve the definition that brings all perspectives together, we encourage researchers to think carefully about the narrative elements that their model focus on, the scope of meaning representation they want to assess, and  phenomena they want to apply their model.

\section*{Acknowledgements}
This research was supported, in part, by the NSF (USA) under Grant Numbers CNS--1948457.

%% file: ijcai22-multiauthor.bbl
\begin{thebibliography}{}

\bibitem[\protect\citeauthoryear{Bajgar \bgroup \em et al.\egroup
  }{2016}]{bajgar2016embracing}
Ondrej Bajgar, Rudolf Kadlec, and Jan Kleindienst.
\newblock Embracing data abundance: Booktest dataset for reading comprehension.
\newblock {\em arXiv preprint arXiv:1610.00956}, 2016.

\bibitem[\protect\citeauthoryear{Bamman \bgroup \em et al.\egroup
  }{2019}]{bamman2019annotated}
David Bamman, Olivia Lewke, and Anya Mansoor.
\newblock An annotated dataset of coreference in english literature.
\newblock {\em arXiv preprint arXiv:1912.01140}, 2019.

\bibitem[\protect\citeauthoryear{Bamman}{2020}]{bamman2020litbank}
David Bamman.
\newblock Litbank: Born-literary natural language processing.
\newblock {\em Computational Humanites, Debates in Digital Humanities (2020,
  preprint)}, 2020.

\bibitem[\protect\citeauthoryear{Brahman \bgroup \em et al.\egroup
  }{2021}]{brahman2021let}
Faeze Brahman, Meng Huang, et~al.
\newblock Let your characters tell their story: A dataset for character-centric
  narrative understanding.
\newblock {\em arXiv preprint arXiv:2109.05438}, 2021.

\bibitem[\protect\citeauthoryear{Brewer}{2017}]{brewer2017literary}
William~F Brewer.
\newblock Literary theory, rhetoric, and stylistics: Implications for
  psychology.
\newblock In {\em Theoretical issues in reading comprehension}, pages 221--240.
  2017.

\bibitem[\protect\citeauthoryear{Britton}{1994}]{britton1994understanding}
Bruce~K Britton.
\newblock Understanding expository text: Building mental structures to induce
  insights.
\newblock 1994.

\bibitem[\protect\citeauthoryear{Chambers and
  Jurafsky}{2008}]{chambers2008unsupervised}
Nathanael Chambers and Dan Jurafsky.
\newblock Unsupervised learning of narrative event chains.
\newblock In {\em Proceedings of ACL 2008}, 2008.

\bibitem[\protect\citeauthoryear{Chambers}{2013}]{chambers2013event}
Nathanael Chambers.
\newblock Event schema induction with a probabilistic entity-driven model.
\newblock In {\em Proceedings of EMNLP 2013}, pages 1797--1807, 2013.

\bibitem[\protect\citeauthoryear{Chen and Choi}{2016}]{chen2016character}
Yu-Hsin Chen and Jinho~D Choi.
\newblock Character identification on multiparty conversation: Identifying
  mentions of characters in tv shows.
\newblock In {\em Proceedings of SIGDIAL 2016}, pages 90--100, 2016.

\bibitem[\protect\citeauthoryear{Chen \bgroup \em et al.\egroup
  }{2016}]{chen2016thorough}
Danqi Chen, Jason Bolton, and Christopher~D Manning.
\newblock A thorough examination of the cnn/daily mail reading comprehension
  task.
\newblock {\em arXiv preprint arXiv:1606.02858}, 2016.

\bibitem[\protect\citeauthoryear{Chen \bgroup \em et al.\egroup
  }{2017}]{chen2017robust}
Henry~Y Chen, Ethan Zhou, and Jinho~D Choi.
\newblock Robust coreference resolution and entity linking on dialogues:
  Character identification on tv show transcripts.
\newblock In {\em Proceedings of CoNLL 2017}, pages 216--225, 2017.

\bibitem[\protect\citeauthoryear{Chen \bgroup \em et al.\egroup
  }{2021}]{chen2021summscreen}
Mingda Chen, Zewei Chu, Sam Wiseman, and Kevin Gimpel.
\newblock Summscreen: A dataset for abstractive screenplay summarization.
\newblock {\em arXiv preprint arXiv:2104.07091}, 2021.

\bibitem[\protect\citeauthoryear{Cot{\'e} \bgroup \em et al.\egroup
  }{1998}]{cote1998students}
Nathalie Cot{\'e}, Susan~R Goldman, and Elizabeth~U Saul.
\newblock Students making sense of informational text: Relations between
  processing and representation.
\newblock {\em Discourse Processes}, 25(1):1--53, 1998.

\bibitem[\protect\citeauthoryear{Delmonte and
  Marchesini}{2017}]{delmonte2017semantically}
Rodolfo Delmonte and Giulia Marchesini.
\newblock A semantically-based computational approach to narrative structure.
\newblock In {\em IWCS 2017}, 2017.

\bibitem[\protect\citeauthoryear{Dunietz \bgroup \em et al.\egroup
  }{2020}]{dunietz2020test}
Jesse Dunietz, Greg Burnham, et~al.
\newblock To test machine comprehension, start by defining comprehension.
\newblock In {\em Proceedings of ACL 2020}, 2020.

\bibitem[\protect\citeauthoryear{Dzendzik \bgroup \em et al.\egroup
  }{2021}]{dzendzik2021english}
Daria Dzendzik, Carl Vogel, and Jennifer Foster.
\newblock English machine reading comprehension datasets: A survey.
\newblock {\em arXiv:2101.10421}, 2021.

\bibitem[\protect\citeauthoryear{Flekova and
  Gurevych}{2015}]{flekova2015personality}
Lucie Flekova and Iryna Gurevych.
\newblock Personality profiling of fictional characters using sense-level links
  between lexical resources.
\newblock In {\em Proceedings of EMNLP 2015}, pages 1805--1816, 2015.

\bibitem[\protect\citeauthoryear{Garner and Bochna}{2004}]{garner2004transfer}
Joanna~K Garner and Cynthia~R Bochna.
\newblock Transfer of a listening comprehension strategy to independent reading
  in first-grade students.
\newblock {\em Early Childhood Education Journal}, 32(2):69--74, 2004.

\bibitem[\protect\citeauthoryear{Glava{\v{s}} \bgroup \em et al.\egroup
  }{2014}]{glavavs2014hieve}
Goran Glava{\v{s}}, Jan {\v{S}}najder, Parisa Kordjamshidi, and Marie-Francine
  Moens.
\newblock Hieve: A corpus for extracting event hierarchies from news stories.
\newblock In {\em Proceedings of 9th LREC}, pages 3678--3683. ELRA, 2014.

\bibitem[\protect\citeauthoryear{Guthrie}{1977}]{guthrie1977research}
John~T Guthrie.
\newblock Research views: Story comprehension.
\newblock {\em The Reading Teacher}, 30(5):574--577, 1977.

\bibitem[\protect\citeauthoryear{Han \bgroup \em et al.\egroup
  }{2021}]{han2021ester}
Rujun Han, I-Hung Hsu, et~al.
\newblock Ester: A machine reading comprehension dataset for reasoning about
  event semantic relations.
\newblock In {\em Proceedings of EMNLP 2021}, pages 7543--7559, 2021.

\bibitem[\protect\citeauthoryear{Hill \bgroup \em et al.\egroup
  }{2015}]{hill2015goldilocks}
Felix Hill, Antoine Bordes, Sumit Chopra, and Jason Weston.
\newblock The goldilocks principle: Reading children's books with explicit
  memory representations.
\newblock {\em arXiv preprint arXiv:1511.02301}, 2015.

\bibitem[\protect\citeauthoryear{Kintsch}{1988}]{kintsch1988role}
Walter Kintsch.
\newblock The role of knowledge in discourse comprehension: A
  construction-integration model.
\newblock {\em Psychological review}, 95(2):163, 1988.

\bibitem[\protect\citeauthoryear{Ko{\v{c}}isk{\`y} and
  others}{2018}]{kovcisky2018narrativeqa}
Tom{\'a}{\v{s}} Ko{\v{c}}isk{\`y} et~al.
\newblock The narrativeqa reading comprehension challenge.
\newblock {\em TACL}, 6:317--328, 2018.

\bibitem[\protect\citeauthoryear{Kry{\'s}ci{\'n}ski \bgroup \em et al.\egroup
  }{2021}]{kryscinski2021booksum}
Wojciech Kry{\'s}ci{\'n}ski, Nazneen Rajani, et~al.
\newblock Booksum: A collection of datasets for long-form narrative
  summarization.
\newblock {\em arXiv preprint arXiv:2105.08209}, 2021.

\bibitem[\protect\citeauthoryear{Labov and Waletzky}{1997}]{labov1997narrative}
William Labov and Joshua Waletzky.
\newblock Narrative analysis: Oral versions of personal experience.
\newblock 1997.

\bibitem[\protect\citeauthoryear{Ladhak \bgroup \em et al.\egroup
  }{2020}]{ladhak2020exploring}
Faisal Ladhak, Bryan Li, Yaser Al-Onaizan, and Kathleen McKeown.
\newblock Exploring content selection in summarization of novel chapters.
\newblock In {\em Proceedings of ACL 2020}, pages 5043--5054, 2020.

\bibitem[\protect\citeauthoryear{Lal \bgroup \em et al.\egroup
  }{2021}]{lal2021tellmewhy}
Yash~Kumar Lal, Nathanael Chambers, Raymond Mooney, and Niranjan
  Balasubramanian.
\newblock Tellmewhy: A dataset for answering why-questions in narratives.
\newblock {\em arXiv preprint arXiv:2106.06132}, 2021.

\bibitem[\protect\citeauthoryear{Levi \bgroup \em et al.\egroup
  }{2020}]{levi2020compres}
Effi Levi, Guy Mor, Shaul Shenhav, and Tamir Sheafer.
\newblock Compres: A dataset for narrative structure in news.
\newblock {\em arXiv preprint arXiv:2007.04874}, 2020.

\bibitem[\protect\citeauthoryear{Liu and others}{2019}]{liu2019neural}
Shanshan Liu et~al.
\newblock Neural machine reading comprehension: Methods and trends.
\newblock {\em Applied Sciences}, 9(18):3698, 2019.

\bibitem[\protect\citeauthoryear{Ma \bgroup \em et al.\egroup
  }{2018}]{ma2018challenging}
Kaixin Ma, Tomasz Jurczyk, and Jinho~D Choi.
\newblock Challenging reading comprehension on daily conversation: Passage
  completion on multiparty dialog.
\newblock In {\em Proceedings of NAACL 2018}, pages 2039--2048, 2018.

\bibitem[\protect\citeauthoryear{Massey \bgroup \em et al.\egroup
  }{2015}]{massey2015annotating}
Philip Massey, Patrick Xia, David Bamman, and Noah~A Smith.
\newblock Annotating character relationships in literary texts.
\newblock {\em arXiv:1512.00728}, 2015.

\bibitem[\protect\citeauthoryear{McNamara and
  Magliano}{2009}]{mcnamara2009toward}
Danielle~S McNamara and Joe Magliano.
\newblock Toward a comprehensive model of comprehension.
\newblock {\em Psychology of learning and motivation}, 51:297--384, 2009.

\bibitem[\protect\citeauthoryear{Mikhalkova and
  others}{2020}]{mikhalkova2020modelling}
Elena Mikhalkova et~al.
\newblock Modelling narrative elements in a short story: A study on annotation
  schemes and guidelines.
\newblock In {\em Proceedings of LREC 2020}, pages 126--132, 2020.

\bibitem[\protect\citeauthoryear{Mirza and Tonelli}{2014}]{mirza2014analysis}
Paramita Mirza and Sara Tonelli.
\newblock An analysis of causality between events and its relation to temporal
  information.
\newblock In {\em Proceedings of COLING 2014}, pages 2097--2106, 2014.

\bibitem[\protect\citeauthoryear{Mostafazadeh \bgroup \em et al.\egroup
  }{2016}]{mostafazadeh2016corpus}
Nasrin Mostafazadeh, Nathanael Chambers, et~al.
\newblock A corpus and cloze evaluation for deeper understanding of commonsense
  stories.
\newblock In {\em Proceedings of the NAACL 2016}, pages 839--849, 2016.

\bibitem[\protect\citeauthoryear{Mou \bgroup \em et al.\egroup
  }{2021}]{mou2021narrative}
Xiangyang Mou, Chenghao Yang, et~al.
\newblock Narrative question answering with cutting-edge open-domain qa
  techniques: A comprehensive study.
\newblock {\em arXiv preprint arXiv:2106.03826}, 2021.

\bibitem[\protect\citeauthoryear{Ning \bgroup \em et al.\egroup
  }{2020}]{ning2020torque}
Qiang Ning, Hao Wu, Rujun Han, Nanyun Peng, Matt Gardner, and Dan Roth.
\newblock Torque: A reading comprehension dataset of temporal ordering
  questions.
\newblock {\em arXiv preprint arXiv:2005.00242}, 2020.

\bibitem[\protect\citeauthoryear{Ostermann \bgroup \em et al.\egroup
  }{2018}]{ostermann2018mcscript}
Simon Ostermann, Ashutosh Modi, Michael Roth, Stefan Thater, and Manfred
  Pinkal.
\newblock Mcscript: A novel dataset for assessing machine comprehension using
  script knowledge.
\newblock {\em arXiv preprint arXiv:1803.05223}, 2018.

\bibitem[\protect\citeauthoryear{Ouyang and McKeown}{2014}]{ouyang2014towards}
Jessica Ouyang and Kathleen McKeown.
\newblock Towards automatic detection of narrative structure.
\newblock In {\em Proceedings of LREC 2014}, 2014.

\bibitem[\protect\citeauthoryear{O’Gorman \bgroup \em et al.\egroup
  }{2016}]{o2016richer}
Tim O’Gorman, Kristin Wright-Bettner, and Martha Palmer.
\newblock Richer event description: Integrating event coreference with
  temporal, causal and bridging annotation.
\newblock In {\em Proceedings of the 2nd Workshop on CNS 2016}, pages 47--56,
  2016.

\bibitem[\protect\citeauthoryear{Papalampidi \bgroup \em et al.\egroup
  }{2019}]{papalampidi2019movie}
Pinelopi Papalampidi, Frank Keller, and Mirella Lapata.
\newblock Movie plot analysis via turning point identification.
\newblock {\em arXiv:1908.10328}, 2019.

\bibitem[\protect\citeauthoryear{Papalampidi \bgroup \em et al.\egroup
  }{2020}]{papalampidi2020screenplay}
Pinelopi Papalampidi, Frank Keller, Lea Frermann, and Mirella Lapata.
\newblock Screenplay summarization using latent narrative structure.
\newblock {\em arXiv preprint arXiv:2004.12727}, 2020.

\bibitem[\protect\citeauthoryear{Paperno \bgroup \em et al.\egroup
  }{2016}]{paperno2016lambada}
Denis Paperno, German Kruszewski, et~al.
\newblock The lambada dataset: Word prediction requiring a broad discourse
  context.
\newblock {\em arXiv:1606.06031}, 2016.

\bibitem[\protect\citeauthoryear{Paris and Paris}{2003}]{paris2003assessing}
Alison~H Paris and Scott~G Paris.
\newblock Assessing narrative comprehension in young children.
\newblock {\em Reading Research Quarterly}, 38(1):36--76, 2003.

\bibitem[\protect\citeauthoryear{Phelan}{1989}]{phelan1989reading}
James Phelan.
\newblock {\em Reading people, reading plots: Character, progression, and the
  interpretation of narrative}.
\newblock University of Chicago Press, 1989.

\bibitem[\protect\citeauthoryear{Piper \bgroup \em et al.\egroup
  }{2021}]{piper2021narrative}
Andrew Piper, Richard~Jean So, and David Bamman.
\newblock Narrative theory for computational narrative understanding.
\newblock In {\em Proceedings of EMNLP 2021}, 2021.

\bibitem[\protect\citeauthoryear{Rashkin \bgroup \em et al.\egroup
  }{2018}]{rashkin2018event2mind}
Hannah Rashkin, Maarten Sap, et~al.
\newblock Event2mind: Commonsense inference on events, intents, and reactions.
\newblock {\em arXiv:1805.06939}, 2018.

\bibitem[\protect\citeauthoryear{Richardson \bgroup \em et al.\egroup
  }{2013}]{richardson2013mctest}
Matthew Richardson, Christopher~JC Burges, and Erin Renshaw.
\newblock Mctest: A challenge dataset for the open-domain machine comprehension
  of text.
\newblock In {\em Proceedings of EMNLP 2013}, 2013.

\bibitem[\protect\citeauthoryear{Rumelhart}{1975}]{rumelhart1975notes}
David~E Rumelhart.
\newblock Notes on a schema for stories.
\newblock In {\em Representation and understanding}, pages 211--236. Elsevier,
  1975.

\bibitem[\protect\citeauthoryear{Ryan}{2007}]{ryan2007toward}
Marie-Laure Ryan.
\newblock Toward a definition of narrative.
\newblock {\em The Cambridge companion to narrative}, 22, 2007.

\bibitem[\protect\citeauthoryear{Saldias and Roy}{2020}]{saldias2020exploring}
Belen Saldias and Deb Roy.
\newblock Exploring aspects of similarity between spoken personal narratives by
  disentangling them into narrative clause types.
\newblock {\em arXiv preprint arXiv:2005.12762}, 2020.

\bibitem[\protect\citeauthoryear{Sang \bgroup \em et al.\egroup
  }{2022}]{sang2022tvshowguess}
Yisi Sang, Xiangyang Mou, Mo~Yu, Shunyu Yao, Jing Li, and Jeffrey Stanton.
\newblock Tvshowguess: Character comprehension in stories as speaker guessing.
\newblock {\em arXiv preprint arXiv:2204.07721}, 2022.

\bibitem[\protect\citeauthoryear{Shank and Abelson}{1977}]{shank1977scripts}
Roger Shank and Robert Abelson.
\newblock Scripts, plans, goals and understanding, 1977.

\bibitem[\protect\citeauthoryear{Sims \bgroup \em et al.\egroup
  }{2019}]{sims2019literary}
Matthew Sims, Jong~Ho Park, and David Bamman.
\newblock Literary event detection.
\newblock In {\em Proceedings of ACL 2019}, pages 3623--3634, 2019.

\bibitem[\protect\citeauthoryear{Walker \bgroup \em et al.\egroup
  }{2006}]{walker2006ace}
Christopher Walker, Stephanie Strassel, et~al.
\newblock Ace 2005 multilingual training corpus.
\newblock {\em LDC, Philadelphia}, 57:45, 2006.

\bibitem[\protect\citeauthoryear{Wolfe and Woodwyk}{2010}]{wolfe2010processing}
Michael~BW Wolfe and Joshua~M Woodwyk.
\newblock Processing and memory of information presented in narrative or
  expository texts.
\newblock {\em British Journal of Educational Psychology}, 80(3):341--362,
  2010.

\bibitem[\protect\citeauthoryear{Xu and others}{2022}]{xu2022fantastic}
Ying Xu et~al.
\newblock Fantastic questions and where to find them: Fairytaleqa--an authentic
  dataset for narrative comprehension.
\newblock {\em arXiv:2203.13947}, 2022.

\bibitem[\protect\citeauthoryear{Yang and Choi}{2019}]{yang2019friendsqa}
Zhengzhe Yang and Jinho~D Choi.
\newblock Friendsqa: Open-domain question answering on tv show transcripts.
\newblock In {\em Proceedings of SIGDIAL 2019}, 2019.

\bibitem[\protect\citeauthoryear{Zan}{1983}]{zan1983toward}
Yigal Zan.
\newblock Toward a functional approach to narrative structure.
\newblock {\em American Anthropologist}, 85(3):649--655, 1983.

\bibitem[\protect\citeauthoryear{Zwaan \bgroup \em et al.\egroup
  }{1995}]{zwaan1995construction}
Rolf~A Zwaan, Mark~C Langston, and Arthur~C Graesser.
\newblock The construction of situation models in narrative comprehension: An
  event-indexing model.
\newblock {\em Psychological science}, 6(5):292--297, 1995.

\end{thebibliography}
